\documentclass[sigconf, nonacm]{acmart}
\usepackage{CJKutf8}
\usepackage{algorithm}
\usepackage{algpseudocode}
\usepackage{amsmath}
\usepackage{multirow}
\usepackage{multicol}
\usepackage{tcolorbox}

\AtBeginDocument{%
  }

\setcopyright{acmlicensed}
\copyrightyear{2018}
\acmYear{2018}
\acmDOI{XXXXXXX.XXXXXXX}

\acmConference[Conference acronym 'XX]{Make sure to enter the correct
  conference title from your rights confirmation emai}{June 03--05,
  2018}{Woodstock, NY}
\acmISBN{978-1-4503-XXXX-X/18/06}

\begin{document}

\title{From Retrieval to Generation: Unifying External and Parametric Knowledge for Medical Question Answering}

\author{Lei Li}
\affiliation{%
  \institution{Gaoling School of Artificial Intelligence, Renmin University of China}
  \city{Beijing}
  \country{China}
}
\email{leil@ruc.edu.cn}

\author{Xiao Zhou}
\authornote{Xiao Zhou and Xian Wu are corresponding authors.}
\affiliation{%
  \institution{Gaoling School of Artificial Intelligence, Renmin University of China}
  \city{Beijing}
  \country{China}
}
\email{xiaozhou@ruc.edu.cn}

\author{Yingying Zhang}
\affiliation{%
  \institution{Tencent Jarvis Lab}
  \city{Beijing}
  \country{China}
}
\email{ninzhang@tencent.com}

\author{Xian Wu\footnotemark[1]}
\affiliation{%
  \institution{Tencent Jarvis Lab}
  \city{Beijing}
  \country{China}
}
\email{kevinxwu@tencent.com}

\renewcommand{\shortauthors}{Li et al.}

\begin{abstract}
Medical question answering (QA) requires extensive access to domain-specific knowledge. A promising direction is to enhance large language models (LLMs) with \emph{external knowledge} retrieved from medical corpora or \emph{parametric knowledge} stored in model parameters. Existing approaches typically fall into two categories: Retrieval-Augmented Generation (RAG), which grounds model reasoning on externally retrieved evidence, and Generation-Augmented Generation (GAG), which depends solely on the model’s internal knowledge to generate contextual documents. However, RAG often suffers from noisy or incomplete retrieval, while GAG is vulnerable to hallucinated or inaccurate information due to unconstrained generation. Both issues can mislead reasoning and undermine answer reliability. To address these challenges, we propose \textbf{\textsc{MedRGAG}}, a unified retrieval--generation augmented framework that seamlessly integrates external and parametric knowledge for medical QA. \textsc{MedRGAG} comprises two key modules: \emph{Knowledge-Guided Context Completion} (KGCC), which directs the generator to produce background documents that complement the missing knowledge revealed by retrieval; and \emph{Knowledge-Aware Document Selection} (KADS), which adaptively selects an optimal combination of retrieved and generated documents to form concise yet comprehensive evidence for answer generation. Extensive experiments on five medical QA benchmarks demonstrate that \textsc{MedRGAG} achieves a 12.5\% improvement over MedRAG and a 4.5\% gain over MedGENIE, highlighting the effectiveness of unifying retrieval and generation for knowledge-intensive reasoning. Our code and data are publicly available at \url{https://anonymous.4open.science/r/MedRGAG}.

\end{abstract}



\maketitle

\section{Introduction}
Large language models (LLMs) have demonstrated remarkable capabilities across a broad spectrum of natural language understanding and question answering tasks~\cite{anil2023palm,touvron2023llama,singhal2023large,zhang2023huatuogpt}. However, their performance remains constrained by the limitations of internal parametric knowledge, which often results in \emph{hallucinations}—plausible yet factually incorrect generations~\cite{ji2023survey,huang2024survey}. These factual inconsistencies pose substantial challenges for medical question answering (QA), a knowledge-intensive task that demands precise reasoning and a high degree of factual reliability. In clinical settings, even minor hallucinations can compromise medical validity, underscoring the necessity for QA systems to generate trustworthy, evidence-grounded responses~\cite{frisoni2022bioreader,zhang2023large,yang2024ensuring}.

\begin{figure}
    \centering
    \includegraphics[width=\linewidth]{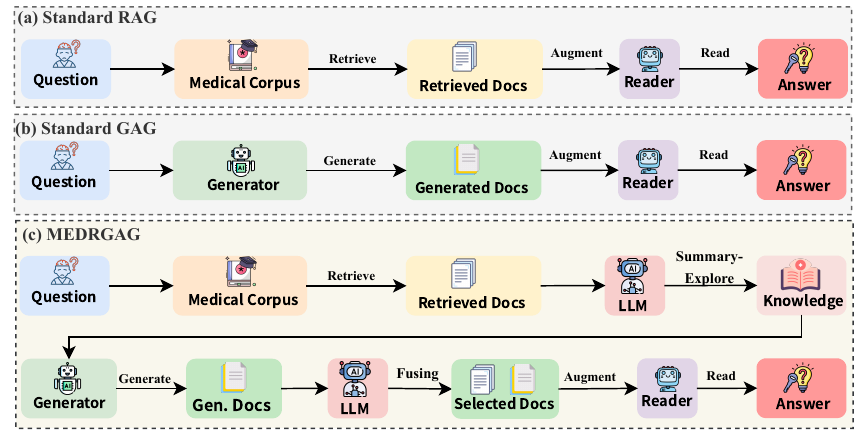}
    \caption{Comparison of \textsc{MedRGAG} with standard RAG and GAG. Figure (a) illustrates RAG, where the retriever extracts external knowledge to enhance the reader’s answer generation. Figure (b) depicts GAG, where the generator activates its internal parametric knowledge to assist the reader in producing answers. Figure (c) presents \textsc{MedRGAG}, which enables the retriever and generator to jointly unify external and parametric knowledge, providing comprehensive evidence to enhance the reader’s answer reliability.}
    \label{fig:intro}
\end{figure}

To enhance the reliability and factual accuracy of LLM-based question answering, recent research has increasingly emphasized \emph{knowledge-augmented generation} methods, which ground model reasoning on external knowledge sources~\cite{zhu2023knowledge,cheng2025survey}. Among these approaches, Retrieval-Augmented Generation (RAG) has emerged as a representative paradigm that follows a \textit{retrieval-then-read} framework (see Figure~\ref{fig:intro} (a)). Specifically, RAG first retrieves relevant information from structured or unstructured medical corpora (e.g., PubMed, Wikipedia, and UMLS) and then conditions the LLM’s answer generation on the retrieved evidence~\cite{asai2024self,xiong2024benchmarking,zhao2025medrag}. By incorporating up-to-date and verifiable documents, RAG effectively grounds the model’s reasoning process, thereby reducing hallucinations and enhancing answer transparency~\cite{li2024enhancing,laban2024summary}. Despite its advantages, the effectiveness of RAG remains constrained by two major factors: (1) retrieved documents are typically chunked into fixed-length passages, often containing noisy or irrelevant information that distracts reasoning~\cite{asai2024self,zhao2025medrag}; and (2) retrieval alone may fail to provide sufficient knowledge coverage, leaving critical information gaps that hinder accurate medical QA~\cite{xiong2024improving,shi2025searchrag}.

In parallel with retrieval-based approaches, another line of research explores \emph{knowledge-augmented generation} by exploiting the \emph{parametric knowledge} embedded within LLMs~\cite{raffel2020exploring,brown2020language}. This paradigm, known as Generation-Augmented Generation (GAG), follows a \textit{generate-then-read} framework (see Figure~\ref{fig:intro} (b)), in which a generator first produces several contextual documents conditioned on the input question, and a reader subsequently leverages these generated contexts to infer the final answer~\cite{yu2022generate,su2023context,frisoni2024generate}. This approach eliminates the reliance on external corpora and enables the model to construct query-specific contexts that more precisely align with the semantic intent of the question. Nevertheless, because GAG relies entirely on generated documents as its knowledge source, it remains susceptible to hallucinated or inaccurate content within those contexts, which can mislead reasoning and ultimately lead to incorrect answers in question answering ~\cite{zhang2023merging,tan2024blinded}.

Recent studies have further explored combining retrieved and generated knowledge to harness the complementary strengths of external and parametric sources~\cite{zhang2023merging,du2025improving}. These approaches primarily aim to mitigate knowledge conflicts that arise during the fusion of retrieved and generated documents, marking a valuable step toward bridging the two paradigms. Nevertheless, fully addressing the intrinsic limitations of each paradigm, including the restricted knowledge coverage of RAG and the potential overreliance on generation-based contexts in GAG, remains a challenging and unresolved problem in knowledge-intensive medical QA.

To address these challenges, we propose \textbf{\textsc{MedRGAG}}, a unified retrieval--generation augmented framework that seamlessly integrates external and parametric knowledge for medical question answering. As illustrated in Figure~\ref{fig:intro} (c), \textsc{MedRGAG} follows a \textit{retrieval–generation–then–read} paradigm: it first employs a retriever to obtain relevant documents from large-scale medical corpora, then leverages a generator to produce complementary background documents, and finally adaptively fuses both sources of evidence, enabling the reader model to generate reliable answers. This design allows \textsc{MedRGAG} to unify retrieval and generation in a coherent pipeline, bridging the gap between external and parametric knowledge sources. Specifically, \textsc{MedRGAG} comprises two core modules:  
(1) \textbf{Knowledge-Guided Context Completion (KGCC)}—this module summarizes and analyzes the retrieved documents to identify missing knowledge required for answering the question, and subsequently guides the generator to produce targeted background documents that complement the retrieved evidence; and  
(2) \textbf{Knowledge-Aware Document Selection (KADS)}—this module groups retrieved and generated documents according to the knowledge requirements of the question and selects a diverse, comprehensive, and non-redundant subset of evidence for downstream reasoning.  
Through the coordinated operation of these two modules, \textsc{MedRGAG} substantially improves knowledge completeness while alleviating overreliance on hallucination-prone generated contexts, thereby enhancing both factual reliability and reasoning robustness in medical QA. 

We conduct extensive experiments on five widely used medical QA benchmarks, including MedQA~\cite{jin2021disease}, MedMCQA~\cite{pal2022medmcqa}, MMLU-Med~\cite{hendrycks2020measuring}, PubMedQA*~\cite{jin2019pubmedqa}, and BioASQ~\cite{tsatsaronis2015overview}. We quantitatively compare performance against representative RAG-based methods such as MedRAG~\cite{xiong2024benchmarking} and GAG-based methods such as MedGENIE~\cite{frisoni2024generate}. To ensure robustness, we evaluate our framework under three different reader architectures: Qwen2.5-7B-Instruct~\cite{team2024qwen2}, LLaMA-3.1-8B-Instruct~\cite{dubey2024llama}, and Ministral-8B-Instruct~\cite{jiang2023mistral7b}. Experimental results demonstrate that \textsc{MedRGAG} achieves an average accuracy gain of 12.5\% over MedRAG and 4.5\% over MedGENIE across all datasets and reader settings. Further analyses show that our framework not only generates more effective complementary background documents but also successfully recovers low-similarity yet highly informative retrieved evidence.
 
In summary, our contributions are threefold:
\begin{itemize}
    \item We propose \textsc{MedRGAG}, a unified retrieval--generation augmented framework that bridges external and parametric knowledge through a coherent \textit{retrieval--generation--then--read} paradigm for medical question answering. 
    \item We design two core modules: Knowledge-Guided Context Completion (KGCC), which generates complementary documents to fill missing knowledge, and Knowledge-Aware Document Selection (KADS), which adaptively selects a reliable combination of retrieved and generated evidence. 
    \item We conduct extensive experiments on five medical QA benchmarks, demonstrating that \textsc{MedRGAG} consistently outperforms a wide range of baselines. Further analyses verify that its two core modules effectively generate complementary background contexts and recover useful evidence.
\end{itemize}

\section{Related Work}
\subsection{Medical Question Answering}
Medical question answering (QA) has long been a fundamental task in biomedical natural language processing~\cite{zweigenbaum2003question,jin2022biomedical}. Early studies primarily rely on BERT-based pretrained models~\cite{devlin2019bert}, which achieve strong performance across various medical benchmarks~\cite{abacha2019overview,lee2020biobert}. With the rapid advancement of large language models (LLMs) such as GPT-4~\cite{achiam2023gpt} and LLaMA~\cite{touvron2023llama}, their exceptional reasoning and comprehension capabilities are also extended to medical QA. To further strengthen domain-specific expertise, a common approach is to continue pretraining these models on large-scale biomedical corpora (e.g., HuatuoGPT~\cite{zhang2023huatuogpt} and PMC-LLaMA~\cite{wu2024pmc}). However, this strategy demands substantial computational resources and high-quality annotated data, which greatly limits its practical applicability. Consequently, knowledge-augmented QA attracts increasing attention as a more flexible and interpretable alternative~\cite{zhu2023knowledge,cheng2025survey}. This paradigm leverages external knowledge as auxiliary reference material, with two representative approaches: Retrieval-Augmented Generation (RAG) and Generation-Augmented Generation (GAG).

\subsection{Medical Retrieval-Augmented Generation}
Retrieval-Augmented Generation (RAG) enhances large language models by grounding their reasoning on external knowledge sources retrieved from large-scale corpora~\cite{lewis2020retrieval,ram2023context}. MedRAG~\cite{xiong2024benchmarking} establishes a comprehensive RAG benchmark and toolkit that integrates hybrid retrieval for medical corpora. Building on this foundation, \textit{i}-MedRAG~\cite{xiong2024improving} introduces iterative follow-up query generation to refine retrieval quality, while Self-BioRAG~\cite{jeong2024improving} incorporates self-reflective retrieval to better handle complex multi-hop medical reasoning. More recently, Omni-RAG~\cite{chen2025towards} proposes a source-planning optimization strategy to retrieve information from diverse medical resources. Despite these advances, RAG-based methods still rely heavily on the relevance and completeness of retrieved documents, which are often noisy and lack critical knowledge. To address these limitations, our framework leverages the intrinsic knowledge of large models to generate complementary background documents and adaptively recover useful evidence while filtering out irrelevant content, thereby constructing a more comprehensive and reliable evidence set for medical QA.

\subsection{Medical Generation-Augmented Generation}
Generation-augmented methods prompt large language models to generate intermediate contexts for question answering, thereby exploiting their internal parametric knowledge~\cite{petroni2019language,roberts2020much}. Representative studies such as GenRead~\cite{yu2022generate} and CGAP~\cite{su2023context} demonstrate that LLMs can serve as strong context generators in open-domain QA. MedGENIE~\cite{frisoni2024generate} applies the \textit{generate-then-read} paradigm to produce multi-view artificial contexts for medical QA. GRG~\cite{abdallah2023generator} and COMBO~\cite{zhang2023merging} further combine generated and retrieved contexts through simple fusion strategies. Despite these advances, these approaches suffer from a critical limitation: the reader often relies heavily on generated documents that may contain hallucinated or inaccurate information. To address this issue, our framework performs multi-step reasoning to generate background documents and integrates trustworthy retrieved evidence to improve the factuality and completeness of the final context, thereby ensuring more reliable answer generation in medical QA.

\section{Methodology}
\subsection{Problem Formulation}

\noindent\textit{\textbf{Definition 3.1.1 (Medical QA).}}  
Given a medical multiple-choice question $q$, which consists of a question stem and an answer set $A = \{a_1, \dots, a_{|A|}\}$, the goal of medical question answering is to identify the most appropriate option $\hat{a}$ as the correct answer.  
In an LLM-based framework, a reader model $\mathcal{M}_r$ takes the question $q$ together with a task-specific prompt $\mathcal{P}_r$ as input and generates the predicted answer:
\begin{equation}
    \hat{a} = \mathcal{M}_r (q, \mathcal{P}_r \mid \theta_r),
\end{equation}
where $\theta_r$ denotes the parameters of the reader model. This basic formulation assumes that the model generates answers solely based on its internal parametric knowledge without external evidence.

\noindent\textit{\textbf{Definition 3.1.2 (RAG).}}  
Retrieval-Augmented Generation (RAG) integrates external knowledge into the input of large language models to enhance their reasoning capability and factual accuracy.  
Formally, given a medical corpus $\mathcal{C} = \{d_1, \dots, d_{|\mathcal{C}|}\}$ and a retriever $\mathcal{R}$ parameterized by $\theta_{\mathcal{R}}$, the retriever identifies the top-$k$ most relevant documents $D_r$ with respect to a question $q$:
\begin{equation}
    D_r = \{d_1^r, \dots, d_k^r\} = \mathcal{R}(q, \mathcal{C} \mid \theta_{\mathcal{R}}).
\end{equation}

The reader model $\mathcal{M}_r$ then takes the question $q$ together with the retrieved document set $D_r$ and a task-specific prompt $\mathcal{P}_r$ as input to generate the predicted answer:
\begin{equation}
    \hat{a} = \mathcal{M}_r (q, D_r, \mathcal{P}_r \mid \theta_r).
\end{equation}

\noindent\textit{\textbf{Definition 3.1.3 (GAG).}}  
In contrast, Generation-Augmented Generation (GAG) employs a large language model as the generator $\mathcal{M}_g$ to produce $k$ tailored background documents $D_g$ for a given question $q$:
\begin{equation}
    D_g = \{d_1^g, \dots, d_k^g\} = \mathcal{M}_g(q, \mathcal{P}_g \mid \theta_g),
\end{equation}
where $\mathcal{P}_g$ denotes the generation prompt designed to encourage diversity and factual reliability, and $\theta_g$ represents the parameters of the generator.  
The reader model $\mathcal{M}_r$ then utilizes these generated documents as auxiliary context to infer the final answer:
\begin{equation}
    \hat{a} = \mathcal{M}_r(q, D_g, \mathcal{P}_r \mid \theta_r).
\end{equation}

\begin{figure*}
    \centering
    \includegraphics[width=\linewidth]{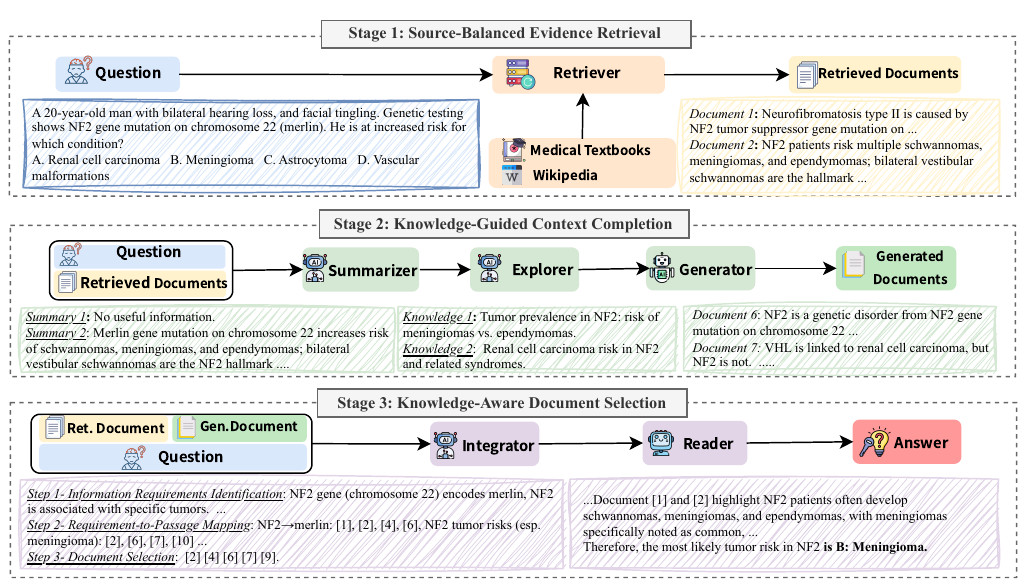}
    \caption{Framework of \texorpdfstring{\textsc{MedRGAG}}{MedRGAG}, which comprises three stages:  source-balanced evidence retrieval, knowledge-guided context completion, and knowledge-aware document selection.}
    \label{fig:whole-model}
\end{figure*}

\subsection{\texorpdfstring{\textsc{MedRGAG}}{MedRGAG}}
We present the overall architecture of our proposed \textsc{MedRGAG} in Figure~\ref{fig:whole-model}, which unifies retrieval and generation to integrate external and parametric knowledge for medical question answering. The framework operates through three sequential stages:  

\begin{itemize}
    \item \textbf{Source-Balanced Evidence Retrieval.}  
    This stage retrieves relevant medical documents from multiple heterogeneous sources. By adopting a source-balanced retrieval strategy, it ensures fair and diverse evidence coverage while reducing the bias toward dominant information sources.  
    \item \textbf{Knowledge-Guided Context Completion.}  
    Building upon the retrieved evidence, this stage identifies missing or incomplete knowledge and generates complementary background documents. Through multi-step reasoning, it enriches the context with accurate and diverse information necessary for answering complex medical questions.  
    \item \textbf{Knowledge-Aware Document Selection.}  
    In the final stage, the framework integrates both retrieved and generated documents and adaptively selects the most relevant and reliable subset as the final evidence for the reader. This selection strikes a balance between factuality and relevance, ensuring the completeness of the supporting knowledge.
\end{itemize}

\subsection{Source-Balanced Evidence Retrieval}
The first stage focuses on retrieving medical documents relevant to the query from multiple heterogeneous knowledge sources. A naïve solution is to merge all sources into a single unified corpus and apply a dense retriever to evaluate query–document similarity, subsequently selecting the top-$k$ documents as evidence. Although such a strategy provides broad coverage, previous studies reveal that retrieving from an overly aggregated corpus can bias the retriever toward dominant data sources or frequently occurring information, thereby compromising the fairness and diversity of the retrieved evidence~\cite{chen2021evaluating,chen2025towards}.  

To address this issue, we employ a \textit{source-balanced retrieval} strategy inspired by Sohn et al.~\cite{sohn2025rationale}.  
Formally, given a question $q$ and a collection of source corpora $\{\mathcal{C}_1, \mathcal{C}_2, \dots, \mathcal{C}_s\}$, the retriever independently selects an equal number of top-ranked documents from each source to construct an intermediate candidate set:
\begin{equation}
\tilde{D}_r = \bigcup_{i=1}^{s} \mathcal{R}(q, \mathcal{C}_i \mid \theta_{\mathcal{R}}),
\end{equation}
thereby ensuring a more balanced and representative distribution of information across different corpora compared with conventional unified-retrieval approaches~\cite{lewis2020retrieval}.

After this balanced retrieval step, the candidate set $\tilde{D}_r$, containing $k \times s$ documents, is subsequently refined through a reranking process.  
We employ MedCPT-Reranker~\cite{jin2023medcpt}, an off-the-shelf cross-encoder model that jointly encodes the question and each candidate document to estimate fine-grained relevance scores. The top-$k$ documents with the highest scores in $\tilde{D}_r$ are then selected to form the final retrieved evidence set $D_r$ for the downstream stages.

\subsection{Knowledge-Guided Context Completion}
While retrieved documents provide valuable external evidence, they still exhibit two major limitations. First, the coverage of medical knowledge is often incomplete: relevant documents may capture only partial aspects of a question and thus fail to deliver comprehensive reasoning support. Second, the retrieved set frequently contains noisy or irrelevant content, which can mislead the reader and ultimately degrade answer accuracy. To address these challenges, we propose a Knowledge-Guided Context Completion (KGCC) module that generates complementary background documents through a three-step process, effectively addressing the problem of insufficient and incomplete knowledge in retrieval-based evidence.

The KGCC process comprises three key steps: summarization of retrieved knowledge, exploration of missing knowledge, and generation of complementary background documents.

\begin{tcolorbox}[title=Summarization Prompt $\mathcal{P}_s$, colback=gray!5!white, colframe=gray!75!black]
You are a professional medical expert. Given the question and a retrieved document, summarize only the useful knowledge points for answering the question. If the document is irrelevant, return "No useful information".
\end{tcolorbox}

\paragraph{Step 1: Summarization of Retrieved Knowledge.}  
We employ a large language model as a summarizer $\mathcal{M}_s$, which extracts and condenses the essential information from each retrieved document with respect to the given medical question. The summarizer generates concise knowledge summaries that retain only the content useful for answering the question while filtering out noisy or irrelevant details. If a document contains no relevant information, the model explicitly outputs “No useful information.” To further improve the quality of summarization, we adopt an in-context learning approach, providing exemplar demonstrations retrieved from the training corpus to guide the model’s summarization behavior.

\paragraph{Step 2: Exploration of Missing Knowledge.}  
Building upon the summarized evidence, we evaluate its adequacy in addressing the medical question. An explorer model $\mathcal{M}_e$ is introduced to identify the most critical missing knowledge points that are necessary yet absent from the current summaries, ensuring that the forthcoming generation stage focuses on complementary information. Formally, the explorer produces a set of missing knowledge items:
\begin{equation}
\mathcal{K} = \{k_1, \dots, k_m\} = \mathcal{M}_e(q, D_s, \mathcal{P}_e \mid \theta_e),
\end{equation}
where $\mathcal{K}$ denotes the set of missing knowledge points and $\mathcal{P}_e$ represents the exploration prompt. Documents labeled as “No useful information” in Step~1 are excluded to improve efficiency.

\begin{tcolorbox}[title=Exploration Prompt $\mathcal{P}_e$, colback=gray!5!white, colframe=gray!75!black]
You are a professional medical expert.  
Given the question and the several pieces of useful information,  
identify the most important missing knowledge required to answer the question thoroughly.
\end{tcolorbox}

\paragraph{Step 3: Generation of Background Documents.}  
Leveraging the missing knowledge points $\mathcal{K}$, the generator $\mathcal{M}_g$ produces background documents that complement the retrieved evidence. Each knowledge point $k_i \in \mathcal{K}$ serves as a guiding signal for generating a corresponding background document:
\begin{equation}
d_i^g = \mathcal{M}_g(q, k_i, \mathcal{P}_g \mid \theta_g),
\end{equation}
This process yields $m$ background documents. When $m < k$, additional documents are generated directly from the question $q$ to ensure a complete set of $k$ background documents $D_g=\{d_1^g,\dots,d_k^g\}$. 

This design provides two major advantages. First, conditioning the generation process on distinct knowledge points encourages diversity among the generated documents. Second, generating supplementary documents directly from the question introduces a broader perspective that enriches the overall background knowledge while avoiding overfitting to localized knowledge.

\subsection{Knowledge-Aware Document Selection}
Given the retrieved and generated documents, the central challenge lies in identifying the most relevant and reliable evidence while eliminating redundant or noisy content. Previous approaches, such as GenRead~\cite{yu2022generate}, simply concatenate the retrieved and generated documents into a single input. While this approach increases knowledge coverage, it also introduces two major drawbacks: (1) irrelevant or low-quality documents are retained, diluting the useful information and distracting the reasoning process; and (2) the enlarged input length substantially increases the contextual load of the reader model, leading to inefficient in reasoning. To address these limitations, we introduce the Knowledge-Aware Document Selection (KADS) module, which adaptively selects a compact yet informative subset of documents, ensuring that the reader model operates on evidence that is both useful and comprehensive.

\begin{algorithm}[b]
\caption{\textsc{MEDRGAG} Framework}
\label{alg:algo}
\begin{algorithmic}[1]
\State \textbf{Input:} Question set $\mathcal{Q}$, Corpus $\mathcal{C}$, Retriever $\mathcal{R}$, Generator $\mathcal{M}_g$, Reader $\mathcal{M}_r$, Summarizer $\mathcal{M}_s$, Explorer $\mathcal{M}_e$, Integrator $\mathcal{M}_i$, Prompts set $\mathcal{P}$ and model parameters $\theta$
\State \textbf{Output:} Predict Answer $\hat{\mathcal{A}}$

\For{each question $q \in \mathcal{Q}$}
\State $D_r = \mathcal{R}(q, \mathcal{C}\mid \theta_{\mathcal{R}})$ \# Retrieve relevant documents

\State $D_s = \mathcal{M}_s(q, D_r, \mathcal{P}_s \mid \theta_s)$ \# Summary useful information

\State $\mathcal{K} = \mathcal{M}_e(q, D_s, \mathcal{P}_e \mid \theta_e)$ \# Explore missing knowledge

\State $D_g = \mathcal{M}_g(q, \mathcal{K}, \mathcal{P}_g \mid \theta_g)$ \# Generate background documents

\State $D_f = \mathcal{M}_i(q, D_r \cup D_g, \mathcal{P}_i \mid \theta_i)$ \# Select documents

\State $\hat{a} = \mathcal{M}_r(q, D_f, \mathcal{P}_r \mid \theta_r)$ \# Produce the final answer

\EndFor

\end{algorithmic}
\end{algorithm}

The proposed knowledge-aware document selection module adaptively integrates retrieved and generated evidence by aligning them with the specific knowledge requirements of each question. A large language model, referred to as the integrator $\mathcal{M}_i$, is prompted to reason over all $2k$ candidate documents and select a compact yet informative subset that best supports answer generation.  
Concretely, $\mathcal{M}_i$ executes the following three reasoning operations:

\begin{itemize}
    \item Knowledge Requirement Identification. The model first analyzes the question to determine the essential knowledge components necessary for a complete and accurate answer.  
    \item Knowledge-to-Document Mapping. Each candidate document is then associated with one or more identified knowledge components based on its content relevance.  
    \item Balanced Evidence Selection. Finally, the model evaluates each knowledge group and selects the top-$k$ documents that collectively maximize knowledge coverage while minimizing redundancy.
\end{itemize}

Considering the strong interdependence among the three reasoning steps, we design a carefully constructed prompt that allows a single LLM to perform the entire selection process in an integrated manner. Formally, the final evidence set $D_f$ is derived as:
\begin{equation}
D_f = \mathcal{M}_i(q, D_r \cup D_g, \mathcal{P}_i \mid \theta_i),
\end{equation}
where $\mathcal{P}_i$ denotes the selection prompt, and $\theta_i$ represents the parameters of the integrator model $\mathcal{M}_i$.  

\begin{tcolorbox}[title=Selection Prompt $\mathcal{P}_i$, colback=gray!5!white, colframe=gray!75!black]
You are a medical expert. Given a question and retrieved documents, select the most useful ones for answering. 

Step 1. Identify key knowledge points required to answer the question.  
Step 2. Map each passage to these knowledge points; assign irrelevant ones to a "No Useful Information" group.  
Step 3. From all groups, select up to 5 passages that ensure coverage and avoid redundancy.
\end{tcolorbox}

\begin{table*}
  \caption{Main experiment results. Best results are in bold and second-best ones are \underline{underlined}.}
  \label{tab:WHOLE}
  \begin{tabular}{l|l|ccccc|l}
  \toprule
    Reader &Method &MedQA-US &MedMCQA &MMLU-Med &PubMedQA* &BioASQ-Y/N &Average \\
     \hline
   \multirow{9}{*}{\centering Qwen2.5-7B}  &Direct Response &62.37 &55.53 &75.76 &37.40 &73.30 &60.87  \\
   & Vanilla RAG &61.12 &57.33 &75.94 &34.20 &72.01 &60.12\\
   & MedRAG &62.61
  & 58.64
 &76.12 &38.00
  & 76.38 &62.35\\
   & \textit{i}-MedRAG  &66.76  &59.12 &76.48 &40.00  & 77.07 &
63.89\\
   & GENREAD  &68.89	&60.24	&78.97	&46.00	&79.29	&66.68\\
   & MedGENIE &69.36 &59.91 &78.60 &\underline{49.20} &81.88 &67.79\\
   &GRG      &69.84 &\underline{60.89} &78.24 &46.60 &81.39 &67.39 \\
   &CGAP     &\underline{72.35} &60.22 &\underline{79.34} &47.80 &\underline{82.36} &\underline{68.41} \\
   &\texorpdfstring{\textsc{MedRGAG}}{MedRGAG}      &\textbf{75.57} &\textbf{62.13} &\textbf{81.63} &\textbf{51.60} &\textbf{84.79} &\textbf{71.14} \\
        \hline
   \multirow{9}{*}{\centering Llama3.1-8B}  &Direct Response &67.48 &58.45 &75.48 &55.40 &76.38 &66.64  \\
   & Vanilla RAG &67.24 &58.38 &75.85 &50.20 &73.30 &64.99\\
   & MedRAG &68.42 & 59.32 & 76.95 &52.00 & 75.24 &66.39
\\
   & \textit{i}-MedRAG  &70.62 &60.63 & 77.31  &53.80  &76.74 & 67.82\\
   & GENREAD  &70.07	&\underline{60.89}	&78.15	&54.60	&78.32	&68.41\\
   & MedGENIE &68.81 &60.24 &78.24 &56.60 &80.26 &68.83\\
   &GRG      &68.97 &\underline{60.89} &76.77 &57.00 &\textbf{82.36} &69.20 \\
   &CGAP     &\underline{71.64} &60.60 &\underline{78.60} &\textbf{58.20} &79.94 &\underline{69.80} \\
   &\texorpdfstring{\textsc{MedRGAG}}{MedRGAG} &\textbf{74.63} &\textbf{61.77} &\textbf{80.90} &\underline{57.80} &\underline{82.04} &\textbf{71.43} \\
        \hline
   \multirow{9}{*}{\centering Ministral-8B}  &Direct Response &57.27 &51.23 &71.63 &31.80 &72.65 &56.92  \\
   & Vanilla RAG &60.64 &54.96 &71.44 &26.20 &72.49 &57.15 \\
   & MedRAG &62.29  & 56.80
 & 76.22
&29.80
  & 74.11
 & 59.84
\\
   & \textit{i}-MedRAG  &66.42  &57.82 & 76.58  &31.20  & 75.59 & 61.53\\
   & GENREAD  &\underline{69.36}	&59.96	&77.04	&41.60	&78.80	&65.35\\
   & MedGENIE &68.03 &59.45 &\underline{78.51} &42.00 &\underline{82.36} &\underline{66.07}\\
   &GRG      &67.24 &\underline{60.08} &77.41 &\underline{42.60} &81.55 &65.78 \\
   &CGAP     &68.42 &60.00 &76.95 &\underline{42.60} &81.07 &65.81  \\
   &\texorpdfstring{\textsc{MedRGAG}}{MedRGAG} &\textbf{74.94} &\textbf{62.13} &\textbf{80.72} &\textbf{44.60} &\textbf{84.14} &\textbf{69.31} \\

\bottomrule
\end{tabular}
\end{table*}
By dynamically balancing retrieved and generated evidence through knowledge-aware grouping, this module effectively reduces noise and redundancy, thereby enhancing reasoning efficiency and answer accuracy.  

The reader model $\mathcal{M}_r$ subsequently takes the refined evidence set as input to generate the final answer:
\begin{equation}
\hat{a} = \mathcal{M}_r(q, D_f, \mathcal{P}_r \mid \theta_r).
\end{equation}

To further clarify the overall workflow, we present the complete \textsc{MedRGAG} framework in pseudo-code, as shown in Algorithm~\ref{alg:algo}.

\section{Experiments}
\subsection{Datasets}
We evaluate \textsc{MedRGAG} on five widely used medical question answering benchmarks: MedQA~\cite{jin2021disease}, MedMCQA~\cite{pal2022medmcqa}, MMLU-Med~\cite{hendrycks2020measuring}, PubMedQA*~\cite{jin2019pubmedqa}, and BioASQ-Y/N~\cite{tsatsaronis2015overview}. All datasets are formulated as multiple-choice QA tasks, where each question includes two to four candidate answers. The overall dataset statistics are summarized in Table~\ref{tab:data_stat}, with additional details presented in Appendix~\ref{sec: appendix-data}. Following standard evaluation practice~\cite{xiong2024benchmarking,frisoni2024generate}, we report accuracy as the primary performance metric.

\begin{table}
  \caption{The statistics of datasets. \#A.: numbers of
options; Avg. L: average token counts in each question.}
  \label{tab:data_stat}
  \small
  \resizebox{0.98\linewidth}{!}{
  \begin{tabular}{l|cccc}
  \toprule
   Dataset &Size &\#A. &Avg. L &Source  \\
     \hline
    MedQA-US &1,273  &4 &177 &Examination \\
   MedMCQA &4,183  &4 &26 &Examination\\
    MMLU-Med &1,089  &4 &63 &Examination\\
    PubMedQA* &500  &3 & 24 & Literature\\
   BioASQ-Y/N & 618  &2 & 17 & Literature\\
\bottomrule
\end{tabular}
}
\end{table} 

\subsection{Baselines}
To comprehensively evaluate the effectiveness of \textsc{MedRGAG}, we compare it with three representative baseline categories. (1) \textbf{Direct response methods}, where the LLM directly answers questions. (2) \textbf{RAG-based methods}, including Vanilla RAG~\cite{lewis2020retrieval}, MedRAG~\cite{xiong2024benchmarking}, and \textit{i}-MedRAG~\cite{xiong2024improving}, which retrieve relevant evidence from external medical corpora. (3) \textbf{GAG-based methods}, such as GenRead~\cite{yu2022generate}, MedGENIE~\cite{frisoni2024generate}, GRG~\cite{abdallah2023generator}, and CGAP~\cite{su2023context}, which synthesize auxiliary contexts from the model’s internal parametric knowledge. A detailed description of each baseline is provided in Appendix~\ref{sec: appendix-baseline}.

In our experiments, we employ BM25~\cite{robertson2009probabilistic} as the retriever, which retrieves candidate documents from multi-corpus: medical textbooks~\cite{jin2021disease} and Wikipedia articles~\cite{xiong2024benchmarking}. We adopt LLaMA-3.1-8B-Instruct~\cite{dubey2024llama} as the generator to produce background contexts. The reader is instantiated with Qwen2.5-7B-Instruct~\cite{team2024qwen2}, LLaMA-3.1-8B-Instruct~\cite{dubey2024llama}, and Ministral-8B-Instruct~\cite{jiang2023mistral7b}. Additionally, we use GPT-4o-mini~\cite{hurst2024gpt} for other LLM-based modules in \textsc{MedRGAG}, including the summarizer, explorer, and integrator. For fair comparison, all retrieval, generation, and fusion stages are standardized to produce a final top-5 set of documents delivered to the reader. Further implementation details are provided in Appendix~\ref{sec: appendix-implementation}.

\subsection{Main Results}
The experimental results are presented in Table~\ref{tab:WHOLE}. Overall, \textsc{MedRGAG} attains the highest average accuracy across all five medical QA benchmarks, consistently outperforming all baseline models. We highlight three key findings below.

\textbf{(1) Effectiveness over direct response methods.} \textsc{MedRGAG} achieves remarkable improvements compared with the direct-response setting, where the reader model answers without any external knowledge. On average, it yields more than a 15\% absolute gain, demonstrating that providing relevant background knowledge significantly enhances answer accuracy. Moreover, this advantage persists across three distinct reader architectures, underscoring the generalizability and robustness of \textsc{MedRGAG}.

\textbf{(2) Superiority to retrieval-augmented methods.} Compared to retrieval-based frameworks such as Vanilla RAG, \textsc{MedRGAG} achieves over 9\% improvement on average across all reader models. Although advanced RAG methods such as MedRAG and \textit{i}-MedRAG enhance retrieval quality via refined selection or iterative querying, their performance gains remain limited. This modest improvement reflects the insufficient knowledge coverage of existing medical corpora, where retrieved documents often fail to encompass all essential medical knowledge. In contrast, \textsc{MedRGAG} leverages its generator to supplement missing knowledge beyond the retrieval corpus, leading to markedly improved answer accuracy.

\textbf{(3) Beyond generation-augmented methods.} \textsc{MedRGAG} also surpasses generation-augmented QA frameworks such as MedGENIE, GenRead, GRG, and CGAP, achieving an average improvement of 4.5\% over MedGENIE. These results show that integrating retrieved evidence into generation-based contexts effectively alleviates the influence of hallucinated documents and yields more accurate reasoning. Furthermore, \textsc{MedRGAG} outperforms GRG, which directly merges retrieved and generated documents, demonstrating that adaptive document selection enables more precise, question-specific evidence aggregation.

\subsection{Ablation Study}

\begin{table}
  \caption{Ablation result on Qwen2.5-7B-Instruct reader.}
  \label{tab:ablation}
  \large
  \resizebox{\linewidth}{!}{
  \begin{tabular}{l|cccc}
  \toprule
   Method &MedQA-US &MedMCQA &MMLU-Med  \\
     \hline
    Direct Response  &62.37 &55.53 &75.76  \\
    \hline
    \texorpdfstring{\textsc{MedRGAG}}{MedRGAG}   &\textbf{75.57} &\textbf{62.13} &\textbf{81.63}\\
   ~~ w/o Generation  &63.47 &58.69 &77.96\\
    ~~ w/o  Retrieval   &72.90 &61.15 &80.35\\
   ~~  w/o KGCC   &72.27	&61.30	&79.43\\
   ~~ w/o KADS  &74.00	&61.10	&80.90\\

\bottomrule
\end{tabular}
}
\end{table}
We conduct ablation studies to assess the contribution of each key component within \textsc{MedRGAG}. The variants are defined as follows: (1) \textit{w/o Generation} removes the generation module and relies solely on retrieved documents for answering. (2) \textit{w/o Retrieval} removes external retrieval and depends only on generated documents. (3) \textit{w/o KGCC} disables the knowledge-guided context completion module, generating background documents directly without identifying missing knowledge. (4) \textit{w/o KADS} removes the knowledge-aware document selection module, instead directly ranking the ten candidate documents to select the top-5 without multi-step reasoning.

As shown in Table~\ref{tab:ablation}, removing either the generation or retrieval component results in a substantial performance drop, indicating that both knowledge sources are indispensable for achieving comprehensive and reliable reasoning. The decline is more pronounced when removing generation, suggesting that generated documents capture question-specific knowledge more effectively than retrieved evidence. Disabling the KGCC module further decreases accuracy, confirming that knowledge-guided completion yields more informative and targeted contexts. Likewise, eliminating the KADS module also degrades performance, demonstrating that adaptive document selection is crucial for identifying useful evidence for each question.

\subsection{Performance Analysis}

\begin{figure}
    \centering
    \includegraphics[width=\linewidth]{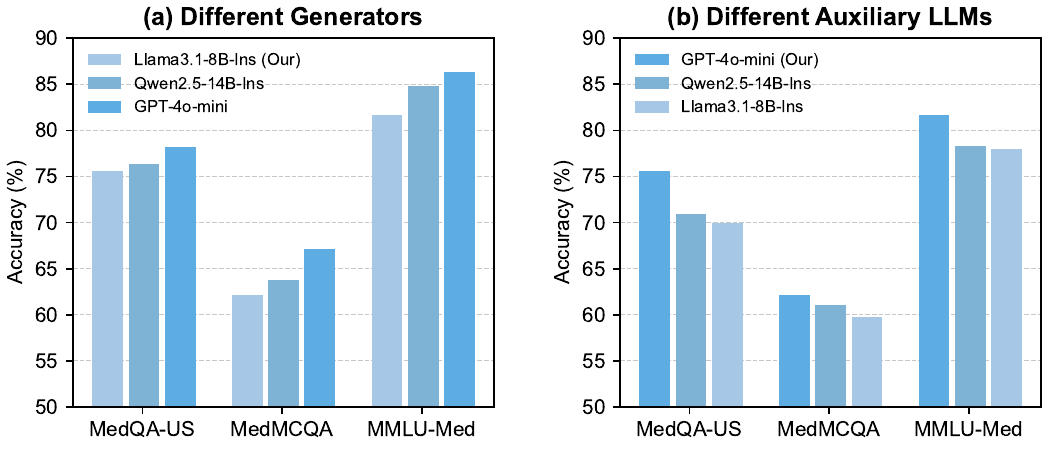}
    \caption{Results of different model scales on the Qwen2.5-7B-Instruct reader. Figures (a) and (b) show different LLM scales used as generators and summarizer-explorer-integrator.}
    \label{fig:performance_analysis}
\end{figure}

\noindent \textbf{Effect of Generator Scale.} We analyze how the generator’s scale influences overall performance by progressively increasing its parameter size from LLaMA3.1-8B-Instruct to Qwen2.5-14B-Instruct and GPT-4o-mini. As illustrated in Figure~\ref{fig:performance_analysis} (a), accuracy consistently rises with larger generator models, indicating that more capable generators can produce higher-quality and broader-coverage knowledge. Compared with the reasoning-intensive clinical dataset MedQA-US, the improvement is particularly pronounced on MedMCQA and MMLU-Med, which focus on fundamental biomedical knowledge, suggesting that larger models possess richer foundational medical understanding.

\noindent \textbf{Effect of Auxiliary LLM Scale.} We further assess how the scale of auxiliary LLMs (serving as the summarizer, explorer, and integrator) impacts overall QA performance. In the main configuration, these components are implemented with GPT-4o-mini, and we additionally evaluate smaller models, including Qwen2.5-14B-Instruct and LLaMA3.1-8B-Instruct. As shown in Figure~\ref{fig:performance_analysis} (b), accuracy declines as the auxiliary model size becomes smaller. The 8B and 14B models perform comparably yet remain substantially below GPT-4o-mini, revealing that smaller LLMs struggle with the complex subtasks of summarization, knowledge exploration, and document selection. Their weaker instruction-following and reasoning capabilities tend to accumulate subtle but compounding errors across stages, ultimately diminishing overall answer accuracy.

\begin{figure}
    \centering
    \includegraphics[width=\linewidth]{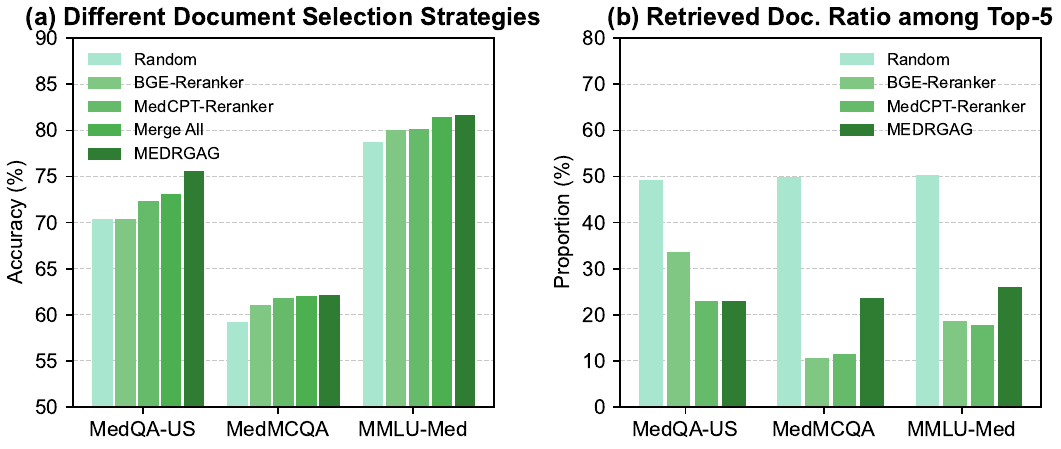}
    \caption{Results of different document selection strategies on the Qwen2.5-7B-Instruct reader. Figure (a) shows different document selection strategies on accuracy. Figure (b) shows the proportion of retrieved documents within the final top-5 across different selection strategies.}
    \label{fig:performance_analysis2}
\end{figure}

\noindent \textbf{Effect of Document Selection Strategies.}
We further analyze how different document selection strategies affect the final answer accuracy. For each question, we use five retrieved and five generated documents produced by \textsc{MedRGAG} and compare several selection methods: \textit{Random}, which randomly samples five documents from ten candidates. BGE-Reranker-Large~\cite{xiao2024c} and MedCPT-Reranker~\cite{jin2023medcpt}, which re-rank all candidates and select the top five documents. \textit{Merge All}, which provides all ten documents to the reader without any filtering. As shown in Figure~\ref{fig:performance_analysis2} (a), our adaptive selection strategy achieves the best overall performance across three datasets. While MedCPT slightly outperforms BGE-Reranker due to its medical-domain optimization, both remain inferior to our approach. Notably, \textsc{MedRGAG} attains comparable or even higher accuracy than the \textit{Merge All} setting while feeding the reader with only half as many documents (5 instead of 10), demonstrating that adaptive document selection effectively retains useful evidence, filters out irrelevant content, and enhances reasoning efficiency.

\noindent \textbf{Analysis of Document Source Preference.}
We further examine how different document selection strategies balance the contributions of retrieved and generated evidence. For each method, we compute the proportion of retrieved documents among the final top-5 evidence, as illustrated in Figure~\ref{fig:performance_analysis2} (b). Overall, generated documents are more favored, aligning with prior findings that ranking models tend to assign higher similarity scores to generative content~\cite{tan2024blinded,frisoni2024generate}. In contrast, our adaptive selection strategy markedly increases the proportion of retrieved documents, suggesting that it successfully recovers valuable retrieval-based evidence that would otherwise be overshadowed by high-similarity generated passages. An exception arises in the MedQA dataset, where our method does not increase the proportion. This is likely because its lengthy and complex question stems cause the retrieved passages to be only partially relevant, rather than directly informative. Overall, these results demonstrate that \textsc{MedRGAG} effectively identifies genuinely useful documents from both sources to enhance QA performance.

\subsection{Case Study}
\begin{table}[t]
\centering
\small
\caption{Case study showing how generated knowledge complements retrieved evidence in \textsc{MedRGAG}.}
\renewcommand{\arraystretch}{1.2}
\begin{tabular}{p{0.95\linewidth}}
\toprule
\textbf{Question:} A 20-year-old man with bilateral hearing loss and facial tingling. Genetic testing shows \colorbox{yellow!30}{NF2 gene mutation} on chromosome 22 (merlin). He is at increased risk for which condition?\\
\textbf{Options:} A. \colorbox{cyan!20}{Renal cell carcinoma} \quad B. \colorbox{green!20}{Meningioma} \quad C. Astrocytoma \quad D. Vascular malformations. \\[3pt]

\textbf{Retrieved Doc. 1:} Neurofibromatosis type II is caused by \colorbox{yellow!30}{NF2 tumor} suppressor gene mutation on chromosome 22. The gene encodes merlin, which normally regulates growth factors... \\
\textbf{Retrieved Doc. 2:}  NF2 patients risk multiple \colorbox{green!20}{meningiomas}, and ependymomas; bilateral vestibular schwannomas are the hallmark. NF2 involves loss of \colorbox{yellow!30}{merlin gene on chromosome 22}... \\
\textbf{Generated Doc. 1:} NF2 is a genetic disorder from \colorbox{yellow!30}{NF2 gene mutation} on chromosome 22. Patients develop bilateral vestibular schwannomas causing hearing loss, and are at increased risk for \colorbox{green!20}{meningiomas}... \\
\textbf{Generated Doc. 2:} VHL is linked to \colorbox{cyan!20}{renal cell carcinoma}, but NF2 is not. NF2 patients develop vestibular schwannomas and may also develop meningiomas and ependymomas... \\
\textbf{Selected Docs:} [Ret\_2], [Gen\_1], [Gen\_2] \\[2pt]
\textbf{Final Answer:} \colorbox{green!20}{B. Meningioma} ~ ~ \textcolor{green!50!black}{\checkmark}  \\
\bottomrule
\end{tabular}
\label{tab:case_nf2}
\end{table}
To further demonstrate the complementary strengths of retrieved and generated knowledge in \textsc{MedRGAG}, we present a representative simplified example in Table~\ref{tab:case_nf2}. For clarity, only the first two retrieved and generated documents are displayed here, while more detailed examples are provided in Appendix~\ref{sec:appendix_case}.  

The question is: \textit{“A 20-year-old male with an NF2 gene mutation is at increased risk for which disease?”} Among the retrieved documents, Doc~1 describes the genetic mechanism of the \textbf{NF2 mutation}, which is relevant but does not specify the associated pathologies, whereas Doc~2 mentions that NF2 patients are predisposed to \textbf{meningiomas} but lacks further explanation. In contrast, the generated documents enrich the reasoning context: Gen~1 elaborates on the etiology of NF2 and its characteristic tumor spectrum, while Gen~2 distinguishes NF2 from \textbf{VHL}-related syndromes, thereby ruling out the distractor option \textit{“renal cell carcinoma.”} Finally, \textsc{MedRGAG} successfully selects the three most informative documents, [Ret\_2], [Gen\_1], and [Gen\_2], to support the reader model in producing the correct answer: \textit{“B. Meningioma”}. This case illustrates how \textsc{MedRGAG} identifies missing knowledge in retrieved evidence, generates complementary documents, and selects the most relevant contexts for accurate reasoning.

\section{Conclusion}
In this paper, we present \textsc{MedRGAG}, a unified retrieval–generation augmented framework that bridges external (retrieved) and parametric (generated) knowledge for medical question answering. Unlike traditional RAG and GAG paradigms that that rely on a single knowledge source, \textsc{MedRGAG} combines both through two key modules: Knowledge-Guided Context Completion (KGCC), which identifies and fills knowledge gaps revealed by retrieval, and Knowledge-Aware Document Selection (KADS), which adaptively selects the most useful evidence for reasoning. Extensive experiments on five benchmark datasets and multiple reader architectures show consistent performance gains over strong retrieval- and generation-based baselines, demonstrating robustness and generalizability. Our analyses and case studies further indicate that \textsc{MedRGAG} mitigates the limitations of incomplete retrieval by recovering valuable retrieval-based evidence while curbing reliance on hallucination-prone generated contexts. In future work, we will extend this framework to broader knowledge-intensive QA tasks and more complex medical scenarios. A promising direction is to unlock large models’ internal knowledge and reasoning ability, integrating it with verifiable retrieved evidence to support trustworthy clinical reasoning.

\bibliographystyle{ACM-Reference-Format}
\bibliography{main}

\newpage
\appendix
\section{Implementation Details}
\label{sec: appendix-implementation}
We construct the retrieval corpus from two sources: a Textbook~\cite{jin2021disease} collection with 125.8K snippets and a Wikipedia~\cite{xiong2024benchmarking} corpus with 29.9M snippets. We employ BM25~\cite{robertson2009probabilistic} as the retriever to retrieve top-32 documents from each source, followed by the MedCPT-Reranker~\cite{jin2023medcpt} to rerank the top-5 most relevant documents.

For each question, three missing knowledge points are identified. The generator, LLaMA3.1-8B-Instruct, operates in a zero-shot setting to produce up to 256 tokens per document. Two additional documents are generated directly conditioned on the question, resulting in five generated contexts. The reader employs a Chain-of-Thought (CoT)~\cite{wei2022chain} reasoning prompt to derive the final answer. We
use GPT-4o-mini for other auxiliary LLMs,
including the summarizer, explorer, and integrator. Temperature is set to 1.2 for the generator and explorer to encourage diversity, and 0.2 for the reader to maintain stability.

All baselines share identical retriever, corpus, and generator configurations for fairness. MedRAG~\cite{xiong2024benchmarking} follows its original setup with four retrievers and Reciprocal Rank Fusion~\cite{cormack2009reciprocal} for aggregation. GenRead~\cite{yu2022generate} retrieves the top-1 document per question in training data and forms five in-context learning (ICL) clusters to generate diverse pseudo-contexts for test data. CGAP~\cite{su2023context} generates five contexts and applies majority voting for the final prediction. GRG~\cite{abdallah2023generator} generates ten candidate documents, from which MedCPT-Reranker selects the top-3 for document fusion.

We adopt the vLLM~\cite{kwon2023efficient} engine for efficient batched inference and memory optimization. GPT-4o-mini is accessed through the OpenAI API for closed-source generation components.

\section{Dataset Details}
\label{sec: appendix-data}
We evaluate \textsc{MEDRGAG} on five representative medical QA datasets covering clinical, biomedical, and professional knowledge domains:

\begin{itemize}
    \item \textbf{MedQA}~\cite{jin2021disease} 
    MedQA is derived from the United States Medical Licensing Examination (USMLE), focusing on diagnosis, treatment, and clinical reasoning. We use the English test subset (1,273 four-choice questions).
    \item \textbf{MedMCQA}~\cite{pal2022medmcqa}
    MedMCQA contains Indian medical entrance-style questions across 21 subjects and 2,400 topics. Since its test labels are unavailable, the official dev set (4,183 questions) is chosen.
    \item \textbf{MMLU-Med}~\cite{hendrycks2020measuring}
    MMLU-Med comprises biomedical questions from six subfields, including Anatomy, College Biology, College Medicine, Clinical Knowledge, Human Genetics, and Professional Medicine, totaling 1,089 test samples.
    \item \textbf{PubMedQA*}~\cite{jin2019pubmedqa} PubMedQA is built from PubMed abstracts with 1,000 expert-annotated questions. To match the RAG setting, we remove original contexts and adopt the 500 official test samples for evaluation.
    \item \textbf{BioASQ-Y/N}~\cite{tsatsaronis2015overview,krithara2023bioasq} 
    We extract all Yes/No questions from machine reading comprehension (Task B) tracks’ gold-standard test sets over the five most recent years (2019–2023), resulting in 618 questions.
\end{itemize}

\section{Baseline Details}
\label{sec: appendix-baseline}
We compare \textsc{MEDRGAG} with a series of representative baselines from both retrieval-augmented (RAG) and generation-augmented (GAG) paradigms:

\begin{itemize}
    \item \textbf{Direct Response} The LLM directly answers each question without external knowledge, evaluating its intrinsic reasoning capability.
    \item \textbf{Vanilla RAG}~\cite{lewis2020retrieval} A standard \textit{retrieval-then-read} framework where a retriever retrieves top-$k$ relevant documents from a corpus using vector similarity, and a reader model subsequently generates an answer.
    \item \textbf{MedRAG}~\cite{xiong2024benchmarking} A medical RAG benchmark combining sparse and dense retrieval to obtain domain-specific evidence from MedCorp, followed by Reciprocal Rank Fusion (RRF)~\cite{cormack2009reciprocal} to enhance retrieval robustness and coverage.  
    \item \textbf{\textit{i}-MedRAG}~\cite{xiong2024improving} An iterative extension of MedRAG that allows the LLM to issue follow-up queries, progressively refining retrieval quality and improving reasoning for multi-hop medical questions.  
    \item \textbf{GenRead}~\cite{yu2022generate} A standard \textit{generate-then-read} framework that first prompts an LLM to synthesize contextual documents based on the input question, and then employs a reader model to generate an answer.
    \item \textbf{MedGENIE}~\cite{frisoni2024generate} The first generation-augmented framework for medical QA, which produces multi-view artificial contexts via option-focused and option-free context to enhance reasoning diversity and factual grounding.  
    \item \textbf{GRG}~\cite{abdallah2023generator} A generator–retriever–generator pipeline where the LLM first generates hypothetical documents, retrieves supporting evidence, and then generates the final answer using both sources.  
    \item \textbf{CGAP}~\cite{su2023context} A two-stage framework performing context generation and answer prediction entirely within a single LLM, emphasizing efficiency and adaptability without relying on external corpora.  
\end{itemize}

We also employ several representative large language models, retrievers, and rerankers in our experiments:

\begin{itemize}
    \item \textbf{Qwen2}~\cite{team2024qwen2} A Mixture-of-Experts model family (0.5B–72B) with strong multilingual reasoning capabilities, pretrained on large-scale corpora and instruction-tuned for general understanding.
    \item \textbf{LLaMA3}~\cite{dubey2024llama} The latest Meta release trained on expanded data with extended context windows, achieving improved coherence and domain adaptability.  
    \item \textbf{Mistral}~\cite{jiang2023mistral7b} An efficient open-weight model employing sliding-window attention for longer context handling with superior quality–efficiency trade-offs.  
    \item \textbf{GPT-4}~\cite{hurst2024gpt} OpenAI’s flagship model trained with reinforcement learning from human feedback (RLHF), offering exceptional reasoning and instruction-following ability.  
    \item \textbf{BM25}~\cite{robertson2009probabilistic} A classical lexical retriever using TF-IDF weighting, implemented with Pyserini to index and retrieve relevant medical snippets.  
    \item \textbf{MedCPT-Reranker}~\cite{jin2023medcpt} A biomedical reranker pre-trained on 255M PubMed click logs, enabling precise semantic matching for domain-specific retrieval.  
    \item \textbf{BGE-Reranker}~\cite{xiao2024c} A general-purpose cross-encoder reranker from BAAI that performs full query–document attention for robust semantic ranking across domains. 
\end{itemize}

\section{Prompts Design}
In this section, we present the detailed prompt templates used in \textsc{MedRGAG}, covering all key components of the framework—including the summarizer, explorer, generator, integrator, and reader.

\begin{tcolorbox}[title=Summarization Prompt $\mathcal{P}_s$, colback=gray!5!white, colframe=gray!75!black]
You are a professional medical expert.  
Given the following question and a retrieved document, distill the useful information that can assist in answering the question.  
Focus only on details directly supported by evidence from the document, and avoid including irrelevant or speculative content.  
If the document does not contain relevant information, return “No useful information.”  
Do not attempt to answer the question—only summarize the essential knowledge needed for answering it accurately.

\medskip
\textbf{Input:}  
Retrieved Document: \{documents\}  
Question: \{question\}  

\textbf{Output:}  
Useful Information: \{useful\_information\}
\end{tcolorbox}

\begin{tcolorbox}[title=Exploration Prompt $\mathcal{P}_e$, colback=gray!5!white, colframe=gray!75!black]
You are a professional medical expert.  
Given the question and several pieces of useful information extracted from retrieved documents, identify the most important missing knowledge required to answer the question thoroughly.  
Analyze the question to determine key knowledge components, compare them with the provided information, and identify the gaps.  
Select the three most critical and non-redundant missing knowledge points, each expressed as a concise conceptual title rather than a full sentence.

\medskip
\textbf{Input:}  
Useful Information: \{information\}  
Question: \{question\}  

\textbf{Output Format:}  
- Reasoning: [Detailed explanation]  
- Knowledge 1: [Conceptual title 1]  
- Knowledge 2: [Conceptual title 2]  
- Knowledge 3: [Conceptual title 3]
\end{tcolorbox}

\begin{tcolorbox}[title=Generation Prompt $\mathcal{P}_g$, colback=gray!5!white, colframe=gray!75!black]
You are a professional medical expert.  
Given the following medical question and a single knowledge point, generate a concise background document that provides relevant explanations or context strictly based on the given knowledge point.  
Do not infer or guess the correct answer, and avoid mentioning any answer options.  
Write in English and keep the content within 256 words.

\medskip
\textbf{Input:}  
Question: \{question\}  
Knowledge Point: \{knowledge\_point\}  

\textbf{Output:}  
Background Document: \{generated\_document\}
\end{tcolorbox}

\section{More Case}
\label{sec:appendix_case}
\begin{table*}[t]
\centering
\small
\caption{Case study showing how generated knowledge complements retrieved evidence in \textsc{MedRGAG}.}
\renewcommand{\arraystretch}{1.2}
\begin{tabular}{p{0.95\linewidth}}
\toprule
\textbf{Example: \#0012 in MedQA}\\[3pt]

\textbf{Question:} A 20-year-old man comes to the physician because of worsening gait unsteadiness and bilateral hearing loss for 1 month. He has had intermittent tingling sensations on both cheeks over this time period. He has no history of serious medical illness and takes no medications. Audiometry shows bilateral sensorineural hearing loss. Genetic evaluation shows a mutation of a \colorbox{yellow!30}{tumor suppressor gene} on chromosome 22 that encodes merlin. This patient is at increased risk for which of the following conditions?"\\
\textbf{Options:} A. \colorbox{cyan!20}{Renal cell carcinoma} \quad B. \colorbox{green!20}{Meningioma} \quad C. Astrocytoma \quad D. Vascular malformations. \\

\textbf{Retrieved Doc. 1:} Neurofibromatosis. Neurofibromatosis type II is caused by a \colorbox{yellow!30}{mutation on chromosome 22}. The mutation falls on the NF2 tumor suppressor gene. The gene normally encodes a cytoplasmic protein known as merlin... The loss of function leads to increased activity of growth factors normally regulated by merlin, leading to the formation of the tumors associated with NF2. \\
\textbf{Retrieved Doc. 2:}  Pathology\_Robbins. Most schwannomas are sporadic, but about 10\% are associated with familial \colorbox{yellow!30}{neurofibromatosis type 2} (NF2). NF2 patients are at risk of developing multiple schwannomas, \colorbox{green!20}{meningiomas, and ependymomas} (the latter are described in Chapter 23). The presence of bilateral vestibular schwannomas is a hallmark of NF2; despite the name, neurofibromas (described later) are not found in NF2 patients... \\

\textbf{Summary Doc. 1:}  No useful information. \\
\textbf{Summary Doc. 2:}  Patients with a mutation of the merlin gene on chromosome 22 are at risk of developing multiple schwannomas, meningiomas, and ependymomas. The presence of bilateral vestibular schwannomas is a hallmark of neurofibromatosis type 2 (NF2). \\

\textbf{Explore Knowledge 1:}  Specific tumor prevalence in NF2 patients (e.g., risk of meningiomas vs. ependymomas). \\
\textbf{Explore Knowledge 2:}  Risk of renal cell carcinoma in NF2 and related syndromes. \\

\textbf{Generated Doc. 6:} **Neurofibromatosis Type 2 (NF2) **\colorbox{yellow!30}{Neurofibromatosis Type 2} (NF2) is a rare genetic disorder characterized by the development of non-cancerous (benign) tumors in the nervous system. The condition is caused by mutations in the NF2 gene, also known as the \colorbox{green!20}{merlin tumor} suppressor gene, which is located on chromosome 22. **Clinical Features:** NF2 patients often present with bilateral vestibular schwannomas (tumors of the nerves responsible for hearing and balance) leading to sensorineural hearing loss... \\
\textbf{Generated Doc. 7:} **Von Hippel-Lindau Syndrome and Neurofibromatosis Type 2 (NF2) Association with \colorbox{cyan!20}{Renal Cell Carcinoma}** Von Hippel-Lindau syndrome (VHL) is a genetic disorder characterized by the development of benign and malignant tumors, particularly hemangioblastomas of the cerebellum, retina, and spinal cord. The NF2 gene encodes for merlin (also known as schwannomin), a tumor suppressor protein that plays a critical role in the regulation of cellular growth and division...\\

\textbf{Selected Docs:} Step 1: Information Requirements Identification: NF2 gene (chromosome 22) encodes merlin, NF2 is associated with specific tumors.

Step 2: Requirement-to-Passage Mapping: NF2→merlin: [1], [2], [4], [6], [9]; NF2 tumor risks (esp. meningioma): [2], [6], [7], [9], [10]; Other tumors (e.g., ependymoma/astrocytoma): [2], [5], [4], [6], [7] Step 3: Document Selection for Completeness and Conciseness: Prioritize NF2→meningioma evidence; ensure RCC exclusion. Final Selection: [2] [4] [6] [7] [9]. \\
\textbf{Final Answer:} \colorbox{green!20}{B. Meningioma} \textcolor{green!50!black}{\checkmark}  \\
\bottomrule
\end{tabular}
\label{tab:more_case}
\end{table*}

\begin{tcolorbox}[title=Selection Prompt $\mathcal{P}_i$, colback=gray!5!white, colframe=gray!75!black]
You are a professional medical expert.  
Given a medical question and ten candidate passages (each labeled with an identifier [id]), select the top-5 most useful passages for answering the question accurately.  

Follow the reasoning steps below:  
1. \textbf{Information Requirements Identification:} Identify the key knowledge points necessary to answer the question thoroughly.  
2. \textbf{Requirement-to-Passage Mapping:} Match each passage to the corresponding knowledge point(s) and classify irrelevant ones into a “No Useful Information” group.  
3. \textbf{Document Selection for Completeness and Conciseness:} Choose up to five passages that together provide comprehensive coverage of the key knowledge points while minimizing redundancy.  

\medskip
\textbf{Input:}  
Documents: \{documents\}  
Question: \{question\}  

\textbf{Output Format:}  
- Reasoning: [Detailed explanation]  
- Final Selection: [id1] [id2] [id3] [id4] [id5]
\end{tcolorbox}

\begin{tcolorbox}[title=Answering Prompt $\mathcal{P}_r$, colback=gray!5!white, colframe=gray!75!black]
You are a professional medical expert.  
Given the question and several retrieved or generated documents, reason step-by-step and provide the final answer.  
First, extract and utilize useful information from the documents; if insufficient, rely on your medical knowledge to complete the reasoning.  
Return your output in JSON format containing both reasoning and the final answer choice.

\medskip
\textbf{Input:}  
Retrieved Documents: \{documents\}  
Question: \{question\}  

\textbf{Output Format:}  
\{"reasoning": "explanation", "answer\_choice": "A/B/C/..."\}
\end{tcolorbox}


\end{document}